\documentclass{article}
\usepackage{spconf,amsmath,graphicx}
\usepackage{booktabs}
\usepackage{arydshln}
\usepackage{color}
\usepackage{mathtools}

\title{Class-Aware Adversarial Lung Nodule Synthesis in CT Images}
\small{
\name{Jie Yang$^{\diamond,\star}$\thanks{$\star$ Equal contribution. This work was conducted when Jie Yang was a research intern at Siemens Healthineers.}, Siqi Liu$^{\dagger,\star}\footnote{*}$, Sasa Grbic$^{\dagger}$, Arnaud Arindra Adiyoso Setio$^{\bullet}$, Zhoubing Xu$^{\dagger}$, Eli Gibson$^{\dagger}$}
\address{\em Guillaume Chabin$^{\dagger}$, Bogdan Georgescu$^{\dagger}$, Andrew F. Laine$^{\diamond}$, Dorin Comaniciu$^{\dagger}$\\
~\\
$^{\diamond}$ Department of Biomedical Engineering, Columbia University, New York, NY, USA \\
$^{\dagger}$ Digital Services, Digital Technology \& Innovation, Siemens Healthineers, Princeton, NJ, USA\\
$^{\bullet}$ Digital Services, Digital Technology \& Innovation, Siemens Healthineers, Erlangen, Germany}
}

\begin{document}

\maketitle

\begin{abstract}
Though large-scale datasets are essential for training deep learning systems, it is expensive to scale up the collection of medical imaging datasets. Synthesizing the objects of interests, such as lung nodules, in medical images based on the distribution of annotated datasets can be helpful for improving the supervised learning tasks, especially when the datasets are limited by size and class balance.
In this paper, we propose the class-aware adversarial synthesis framework to synthesize lung nodules in CT images.
The framework is built with a coarse-to-fine patch in-painter (generator) and two class-aware discriminators.
By conditioning on the random latent variables and the target nodule labels, the trained networks are able to generate diverse nodules given the same context.
By evaluating on the public LIDC-IDRI dataset, we demonstrate an example application of the proposed framework for improving the accuracy of the lung nodule malignancy estimation as a binary classification problem, which is important in the lung screening scenario. We show that combining the real image patches and the synthetic lung nodules in the training set can improve the mean AUC classification score across different network architectures by $2\%$.
\end{abstract}
\section{Introduction}
\label{sec:intro}
\begin{figure*}
    \centering
    \includegraphics[width=0.9\textwidth]{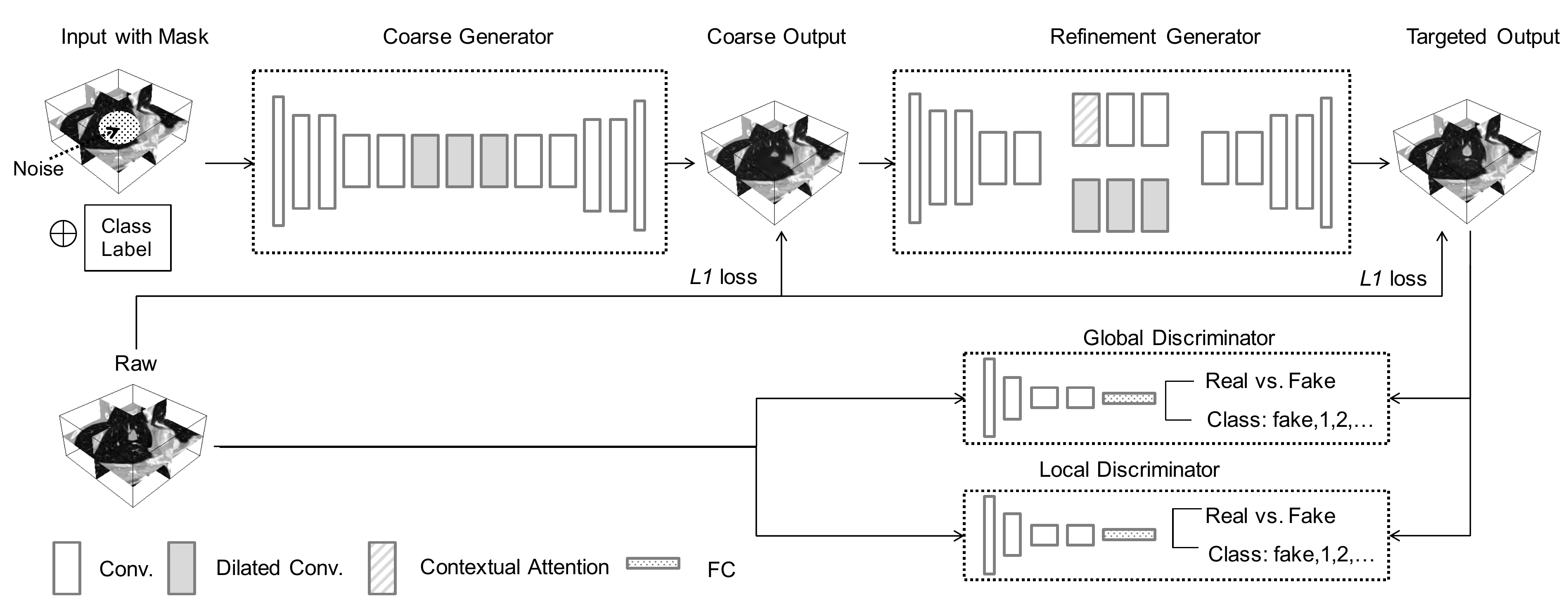}
    \caption{The proposed learning framework. The noise masked 3D image patch $x_{masked}$ along with the class label $c$ is fed into a sequence of two generator networks. The raw patch is used for computing the reconstruction $L1$ loss for both the coarse generator $G_1$ and the refinement generator $G_2$. Besides the $L1$ loss, the generators are also optimized with the adversarial losses provided by both the local discriminator $D_{local}$ and the global discriminator $D_{global}$. Each discriminator is responsible for predicting if a patch is fake as well as the nodule malignancy label as the auxiliary classification output.}
\label{fig:framework}
\end{figure*}

Deep convolutional neural networks (CNN) have demonstrated superior performance for medical image analysis tasks, for example, image segmentation, object detection, and classification. 
The performance of such methods is however constrained by the quantity and the diversity of the training data available. 
Given the imbalanced distribution in the real-world data, the rare and abnormal cases are generally underrepresented in the training data. 
It is thus important to improve the data efficiency of the medical machine learning systems by either improving the supervised learning approaches or synthesizing images based on the existing annotated data. 
This paper explores the latter path to approach this problem, with specific application to classify benign versus malignant lung nodules in CT images.
Basic data augmentation techniques, such as random cropping, shifting, scaling, flipping and rotations, can be used to introduce a certain level of diversity during training stage, but cannot account for the diversity of nodule morphology and locations. 
Some recent studies proposed to use generative adversarial network (GAN) networks to synthesize lesions in medical image patches to augment the training data \cite{wu2018conditional,jin2018ct}. Such methods train the GAN networks to in-paint a cropped area with the objects of interests, such as lesions. 
The generator network is trained with a reconstruction loss between the synthetic patch and the real patch as well as an adversarial loss produced by a discriminator network.
They concluded that using synthetic patches could improve the performance of supervised learning tasks.
However, such networks were designed to generate objects conditioning based on only the surrounding context and random noises, lacking the capability of generating objects with manipulable properties which we believe to be important for many machine learning applications in medical imaging, such as balancing the datasets for classification tasks. In a more recent study, authors propose to use synthetic shape models to condition the nodule synthesis \cite{liu2018}.

In this paper, we propose an adversarial learning framework to synthesize lung nodules in CT images by conditioning on the target categories and sizes.
Hence the nodules can be in-painted at random locations with manipulable attributes.
We demonstrate an example application of the class-aware synthesized nodules, which is to augment training images in an unbalanced dataset for improving the assessment of the nodule malignancy risk.
A data-driven model that can accurately predict nodule malignancy risk from CT images may improve management decisions and prevent unnecessary imaging or invasive follow-up procedures on benign nodules, thus increase the effectiveness of lung cancer screening programs \cite{MacMahon2005GuidelinesSociety}.
By evaluating on the public Lung Image Database Consortium (LIDC-IDRI) dataset \cite{Arma15,Arma11,Clar13}, we show that our synthesized nodule patches are beneficial for improving the performance on estimating the nodule malignancy risk.
The proposed framework has the potential to be generalized to synthesizing other objects of interests in medical images. 

\section{Methods}
The proposed framework to generate synthetic nodules is formulated as an in-painting problem which fills a masked area in a 3D lung CT image with a pulmonary nodule with the specified category and size.
The framework contains two major components: 1) two generators to perform coarse-to-fine in-painting by incorporating contextual information; 
2) local and global discriminators to enforce the local quality and the global consistency of the generated patches, and auxiliary domain classifiers to constrain the generated nodules with the manipulable properties, such as the malignancy. The overall framework is depicted in Fig.~\ref{fig:framework}.

\subsection{Coarse-to-Fine Reconstruction}

The 3D input nodule patches are extracted from the lung CT images centering on annotated nodules. 
The nodule in each patch is replaced with a 3D spherical noise mask generated according to the annotated nodule diameter. 
The masked patch and a class-label map are firstly fed into a 3D hour-glass CNN $G_1$, as shown in Fig.~\ref{fig:framework}. $G_1$ is designed to be easier optimized and can reconstruct a coarsely synthesized nodule in the masked region to facilitate subsequent learning. 
The output of $G_1$ is fed into another network $G_2$ which has a similar architecture as $G_1$, to refine the details in the output map. $G_1$ and $G_2$ together form a stacked image in-painting generator $G$. Both $G_1$ and $G_2$ are optimized with the reconstruction loss $L_{recon}$ between the reconstructed patch and the real nodule patch 
\begin{equation}
    L^{(1,2)}_{recon}=L_{masked} + \lambda_{1} L_{global}
\end{equation} 
where $L_{masked}$ and $L_{global}$ are the normalized $L1$ loss across the masked area and the entire patch, respectively.
By optimizing $L_{recon}$, the stacked generator $G$ is trained to reconstruct the nodules in the original patch based on the lung tissue context, random noise mask and the class label.
Aside from $L_{recon}$, $G_2$ is also optimized by an adversarial loss provided by two discriminator networks described in Sec.~\ref{sec:class-aware}.

Features from surrounding regions can be helpful for in-painting the boundary of the nodules in the masked area. In a recent study \cite{Yu2018GenerativeAttention}, a contextual attention model is proposed to borrow the textures from the background patches to generate missing patches. 
We use the contextual attention model to match between the foreground and background textures by measuring the normalized cosine similarity of their features.
An attention map on the background voxels is obtained for reconstructing foreground area with the attention filtered background features. 
The contextual attention layer is differentiable and fully-convolutional. 
We refer the complete definition the contextual attention to the original paper for brevity.

\subsection{Class-Aware Synthesis}
\label{sec:class-aware}
Two discriminator networks $D_{local}$ and $D_{global}$ are used to optimize $G_1$ and $G_2$ in an adversarial approach together with the reconstruction losses. 
$D_{local}$ is applied to the masked area only to improve the nodule appearance.
$D_{global}$ is applied to the entire patch for the global consistency of the in-painting.
We denote both discriminators as $D_{*}$ here for brevity. 
We use the conditional Wasserstein GAN objective and enforcing the gradient penalty \cite{Gulrajani2017ImprovedGANs} to train the local and global discriminators $D_{*}$ and the stacked generator $G$ as 
\begin{equation}
    \begin{multlined}
    L_{adv} = E_x[D_{*}(x)] - E_{z,c}[D_{*}(G(x_{masked},z,c))] \\
    - \lambda_{gp} E_{\hat{x}} [(\|\nabla D_{*}(\hat{x})\|_2 - 1)^2]
    \end{multlined}
\end{equation}
where $x$ is sampled from a real patch, $G(x_{masked},z,c)$ is the generator in-painting output, $\hat{x}$ is sampled uniformly between real patches and generated patches.
The class label $c$ is replicated to the same size as the input patch $x$ and is concatenated with $x$ as another input channel.
Besides the WGAN discriminators $D_*$, we add an auxiliary domain classifier $D_{cls}$ on the top of each discriminator network to ensure that $G$ generates nodules in the targeted class $c$.
In this training objective, each $D_{cls}$ attempts to classify the output patch into the domain class $c'$ (0 = fake, 1 = benign, and 2 = malignant). 
The label 0 is used to prevent the generator from duplicating nodules that are easy to classify but less diversified. 
$D_{cls}$ is optimized with the class-aware loss as
\begin{equation}
    L_{cls} = E_{x,c'}[-logD_{cls}(c'|x)]
\end{equation}
where $D_{cls}(c'|x)$ represents a probability distribution over domain classes $c'$.
Though both $D_{*}$ and $D_{cls}$ are optimized to discriminate fake and real patches, empirically we find it hard for the learning system to converge without $L_{adv}$.
In practice, $L_{cls}$ is added after the generator $G$ is well-trained to in-paint real-look nodules.
In this adversarial learning problem, $G$ tries to in-paint the patch that can be classified as the target domain $c$ as well as to fool $D_{*}$ to misjudge them in the distribution of the real patches. 
The objective for the whole class-aware nodule synthesis learning can then be summarized as
\begin{equation}
    L_{D_*} = L_{adv} + \lambda_{cls}^{(D_*)} L_{cls}
\end{equation}
\begin{equation}
    L_{G} = - L_{adv} + \lambda_{cls}^{(G)} L_{cls} + \lambda_{recon}L_{recon}
\end{equation}

\subsection{Application in 3D Nodule Malignancy Classification using Video Pre-trained Networks}
In the context of the action recognition tasks in videos, the spatiotemporal 3D CNNs are recently demonstrated as more effective than the 2D CNNs when large-scale frame-wise annotated datasets such as the Kinetics is available \cite{Hara2017CanImageNet}. 
To classify the CT nodule patches into benign and malignant classes, we use 3D CNNs pre-trained on natural video classifications \cite{Hara2017CanImageNet}. 
The temporal dimension of video data resembles the depth dimension of 3D medical image volumes. 
It is also easier to scale up the collection of video datasets than the medical imaging datasets.
Thus, using video pre-trained networks could be helpful for stabilizing the network training and preventing over-fitting.

\begin{figure*}[t]
    \centering
    \includegraphics[width=.9\linewidth]{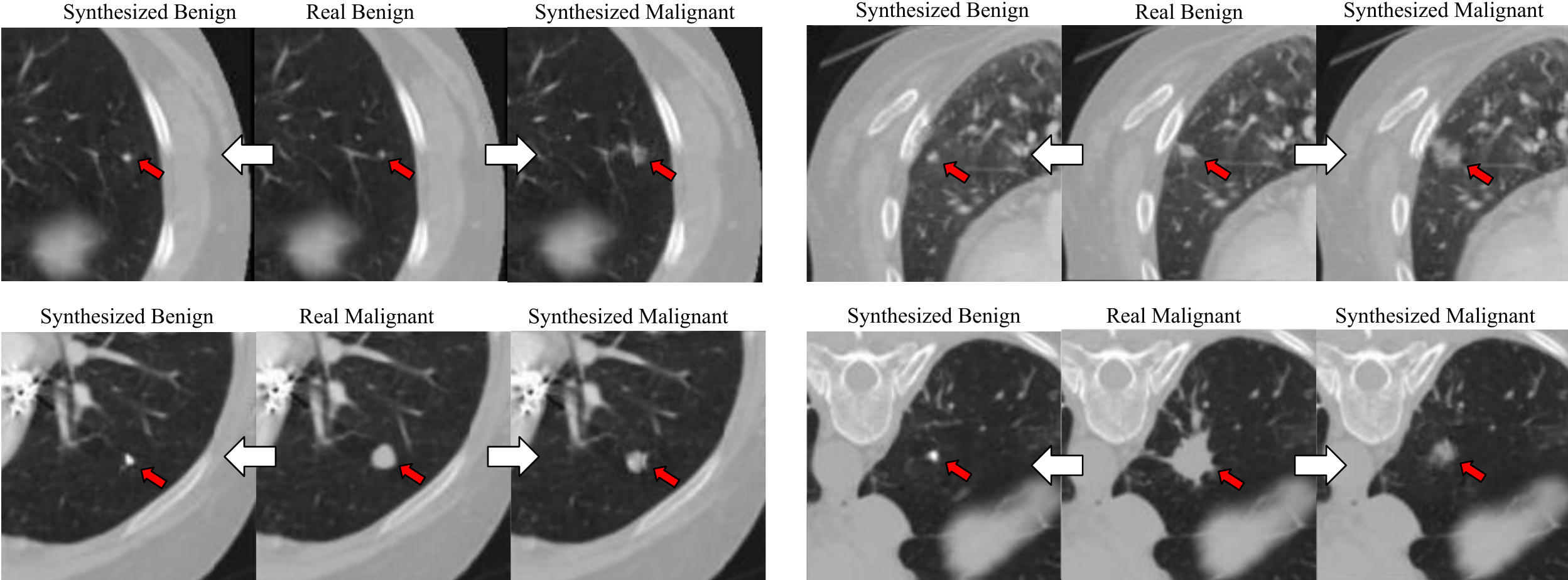}
    \caption{Example outputs of the nodule synthesis by altering the nodule malignancy label to benign or malignant. The generated benign nodules tend to have smaller sizes as well as clearly defined boundary.}
    \label{fig:label_conversion}
\end{figure*}

\begin{figure}
    \centering
    \includegraphics[width=.9\linewidth]{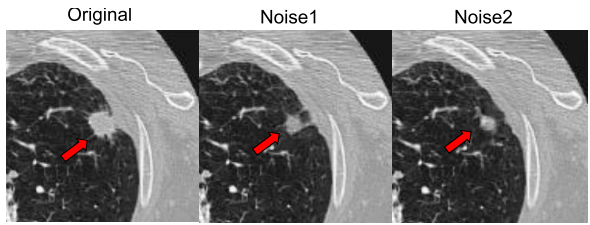}
    \caption{Example outputs of the nodule synthesis with the same input patch and different the initial noise masks.}
    \label{fig:noise}
\end{figure}

\section{Results}

We evaluated the proposed methods on the public The Lung Image Database Consortium (LIDC-IDRI) dataset \cite{Arma15,Arma11,Clar13} consisting of diagnostic and lung cancer screening thoracic computed tomography (CT) scans with annotated lesions. 
The LIDC-IDRI dataset consists of 1,010 patients and 1,308 chest CT imaging studies in total. The nodules in LIDC-IDRI were annotated by four radiologists. The likelihood of malignancy of each nodule is assessed, and a score ranging from 1 (highly unlikely) to 5 (highly suspicious) is given by each radiologist.
We define the nodules with the majority score $\geq 4$ to be malignant and the rest to be benign.
In our experiments, we extracted the nodule patches from the LIDC dataset with the resolution $1\times1\times2$ mm ($64\times64\times32$ voxels). 
The patches were randomly split into the training set, validation set and testing set according to the patients as shown in Table~\ref{tbl:data-split}.

\begin{table}[b]
\centering
\resizebox{0.7\columnwidth}{!}{
\begin{tabular}{@{}llll@{}}
\toprule
Subset     & Benign ($< 4$) & Malignant ($\geq4$) & Total \\ \midrule
Train      & 1016   & 233       & 1249  \\
Train+Syn  & 1016   & 696       & 1712  \\
Validation & 128    & 28        & 156   \\
Test       & 133    & 24        & 157   \\ \bottomrule
\end{tabular}
}
\caption{The breakdown of train, validation and test patch partition.}
\label{tbl:data-split}
\end{table}

\begin{table}[]
\centering
\resizebox{0.9\columnwidth}{!}{
\begin{tabular}{@{}lllll@{}}\toprule
\textbf{Network}             & \textbf{ACC}    & \textbf{SEN}    & \textbf{SPE}    & \textbf{AUC}    \\ \midrule
Raw                 &        &        &        &        \\ \midrule
ResNet-50           & 0.8216 & 0.6666 & 0.8496 & 0.8286 \\ 
ResNet-101          & 0.8853 & 0.6666 & 0.9248 & 0.8320  \\ 
ResNet-152          & 0.8917 & 0.5833 & 0.9473 & 0.8756 \\ 
ResNext-101         & 0.9044 & 0.6249 & 0.9548 & 0.8558 \\ 
Mean                & 0.8757 & 0.6353  & 0.9191 & 0.8480   \\ \midrule
Raw + Weighted Loss &        &        &        &        \\ \midrule
ResNet-50           & 0.8662 & 0.6666 & 0.9022 & 0.8161 \\ 
ResNet-101          & 0.7961 & \textbf{0.7499} & 0.8045 & 0.8154 \\ 
ResNet-152          & 0.8598 & 0.7083 & 0.8872 & 0.8098 \\ 
ResNext-101         & 0.8471 & 0.6666 & 0.8796 & 0.8342 \\ 
Mean                & 0.8423  & 0.6978  & 0.8683 & 0.8188 \\ \midrule
Raw + Synthesis           &        &        &        &        \\ \midrule
ResNet-50           & 0.8216 & 0.7083 & 0.8421 & 0.8746 \\ 
ResNet-101          & 0.9044 & 0.5833 & 0.9624 & 0.8458 \\ 
ResNet-152          & \textbf{0.9171} & 0.5833 & \textbf{0.9774} & \textbf{0.8812} \\ 
ResNext-101         & 0.8853 & 0.6666 & 0.9248 & 0.8768 \\ 
Mean                & 0.8821  & 0.6353 & 0.9266 & 0.8696   \\ \bottomrule
\end{tabular}
}
\caption{The nodule malignancy classification results of different network architectures and different data balancing strategies.}
\label{tbl:numbers}
\end{table}

We trained the proposed nodule synthesis framework on the training patches only. 
In Fig.~\ref{fig:label_conversion}, we demonstrate the nodule synthesis results by conditioning on different labels. 
Given the same background patch, the framework is capable of generating nodules with different specified malignancy labels.
We also show in Fig.~\ref{fig:noise} that the network could generate nodules with different morphology and textures using different noise masks.
The trained generator is used for synthesizing 463 patches containing malignant patches from malignant patches randomly sampled from the real training malignant patches since malignant nodules are relatively rare in the original LIDC-IDRI dataset.
The synthetic nodule patches are combined with the original training patches to train the 3D classification CNNs.
To evaluate the effectiveness of using the synthetic patches on estimating the lung nodules malignancy, we trained four 3D CNN architectures with different capacities: ResNet-50, ResNet-101, ResNet-152 \cite{He2016}, and ResNext-101 \cite{Xie2017AggregatedNetworks}.
All the networks were initialized with the weights pre-trained on the Kinetic video dataset \cite{Hara2017CanImageNet,Kay2017TheDataset}.
The cross-entropy loss was used for training the CNN classifiers. 
We also evaluated the differences between the unweighted (Raw) and weighted cross entropy loss (Raw + Weighted Loss) with the weights accounting for training sample class distribution.
Traditional data augmentation methods including random cropping and scaling were used for training all the networks.
The testing accuracy (ACC), sensitivity (SEN), specificity (SPE), and the area under the ROC curve (AUC) are presented in Table.~\ref{tbl:numbers}. The models were selected based on the highest AUCs on the validation set. With the synthetic patches (Raw + Synthesis), the 3D ResNet152 achieved the highest accuracy, specificity, and AUC score across all the experiments. Though the weighted loss is typically used in the imbalanced dataset, it did not show better AUC than neither the unweighted loss nor the augmented dataset.
The increase of the mean of the AUC scores also indicates that using the synthetic nodule patches are helpful for improving the nodule malignancy classification performance.

\section{Conclusions}
In this paper, we propose an adversarial in-painting based framework for synthesizing lung nodules with class-aware manipulations. We demonstrate one example application of the generated lung nodule patches on the classification of nodule malignancy. The qualitative results show that the proposed framework is capable of generating lung nodules with the specified malignancy labels. By evaluating on the nodule patches obtained from the LIDC-IDRI dataset, we show that the generated nodules can be helpful for improving the classification performance on an imbalanced dataset.

\noindent\textbf{Disclaimer}: This feature is based on research, and is not commercially available. Due to regulatory reasons, its future availability cannot be guaranteed.
\bibliographystyle{IEEEbib}
\bibliography{isbi19_nodule.bib}

\end{document}